\documentclass[conference]{IEEEtran}

\def\e{\epsilon}
\usepackage{authblk}
\usepackage{caption}
\usepackage{amsmath}
\usepackage{amssymb}
\usepackage{url}
\usepackage{graphicx}
\usepackage{rotating}
\usepackage{ifthen}
\usepackage{graphicx}
\usepackage{epsfig}
\usepackage{multirow}
\usepackage{array}
\usepackage{color}
\usepackage{fancybox}
\usepackage{blindtext}
\usepackage{soul}
\usepackage{flushend}

\usepackage{algorithm,algorithmic}
\usepackage{xcolor}
\newcommand{\memo}[2]{\textcolor{#1}{#2}}

\renewcommand{\memo}[2]{} 

\def\argmin{\operatornamewithlimits{argmin}}
\def\argmax{\operatornamewithlimits{argmax}}

\newfloat{subroutine}{htbp}{loa}
\floatname{subroutine}{Subroutine}
\makeatletter\newcommand\l@subroutine{\@dottedtocline{1}{1.5em}{2.3em}}\makeatother

\DeclareCaptionType{copyrightbox}

    \def\E{{\mathbb E}}

    \title{Learning opening books in partially observable games:\\using random seeds in Phantom Go}

\author[1]{Tristan Cazenave}
\author[2,3]{Jialin Liu}
\author[4]{Fabien Teytaud}
\author[2]{Olivier Teytaud}
\affil[1]{Lamsade, Univ. Paris Dauphine, Paris, France \authorcr Email: {\tt cazenave@lamsade.dauphine.fr}}
\affil[2]{Inria, CNRS UMR 8623, Univ. Paris-Sud, Gif-sur-Yvette, France \authorcr Email: {\tt \{lastname.firstname\}@inria.fr}}
\affil[3]{CSEE, Univ. of Essex, Colchester, UK \authorcr Email: {\tt jialin.liu@essex.ac.uk}}
\affil[4]{Lisic, Univ. Littoral, France, Calais, France \authorcr Email: {\tt fabien.teytaud@lisic.univ-littoral.fr}}
    
\begin{document}
    \maketitle

    \begin{abstract}
	    Many artificial intelligences (AIs) are randomized. One can be lucky or unlucky with the random seed; we quantify this effect and show that, maybe contrarily to intuition, this is far from being negligible. Then, we apply two different existing algorithms for selecting good seeds and good probability distributions over seeds. This mainly leads to learning an opening book. We apply this to Phantom Go, which, as all phantom games, is hard for opening book learning. We improve the winning rate from 50\% to 70\% in 5x5 against the same AI, and from approximately 0\% to 40\% in 5x5, 7x7 and 9x9 against a stronger (learning) opponent.
    \end{abstract}


\section{Introduction}

\subsection{Offline learning in games}
Offline learning in games can be e.g. endgame table building \cite{nalimov}, opening book construction by self-play \cite{gaudel2010principled}, or parameter estimation \cite{chaslot08b}.
We propose the use of Random-Seed-portfolios, which consists in optimizing the probability distribution on random seeds, for offline learning in games for which randomized AIs perform well. This approach will essentially, though not only and not explicitly, learn at the level of the opening book. Learning opening books is particularly hard in partially observable games, due to the difficult belief state estimation; therefore, this recent approach by random seeds is particularly suitable in this case. 

The random seeds approach has already been proposed for the game of Go \cite{cazenave2015rectangular,st2015nash}, but the present paper is, to the best of our knowledge, the first application to \emph{partially observable games}, and our resulting algorithm outperforms by far the original algorithm, which is at the current top level in Phantom Go, including its traditional board sizes. This is mainly obtained through opening book learning - which is a hard task in partially observable games.


\subsection{Randomized artificial intelligences}\label{randai}

\subsubsection{Why randomizing AIs.}
There are games in which optimal policies are randomized and, beyond that, in many cases the state of the art is made of randomized algorithms, in particular since the advent of Monte Carlo Tree Search \cite{coulom06,uct}. Randomized AIs are also required when the AI should be robust to ``overfitting'' by an opponent - i.e. when we do not want an opponent to be able to learn, by repeated games, a simple winning strategy. A deterministic AI is certainly not suitable in such a case, e.g. for playing on a server or for the pleasure/education of a human opponent. Still, we point out that our approach makes sense in terms of pure performance against the baseline algorithms.

\subsubsection{The original Monte Carlo approach in games.}
The Monte Carlo approach in games goes back to \cite{bruegmann}. The basic idea is to evaluate a position using random simulations. The value at a state $s$ is obtained by averaging the result of hundreds of games played randomly from this state $s$. This is compliant with partially observable games by randomly sampling the hidden parts of the state. With ad hoc randomization, this approach is the state of the art in Phantom Go \cite{cazenave2005phantom}.

\subsubsection{Improvements of the original Monte Carlo approach.}
The original Monte Carlo method for games has been vastly improved \cite{bouzy02,bouzy03}.
For fully observable games it was outperformed by Monte Carlo Tree Search \cite{coulom06}, which adds a tree search to the Monte Carlo evaluation principle. For fully observable puzzles (one player games), nested Monte Carlo often outperforms Monte Carlo \cite{nestedMC,nesteduct}. In partially observable games with large number of hidden states, Monte Carlo remains at the top of game programming \cite{phantomgo,phantomgo2}.

\subsection{Boosting randomized artificial intelligences and learning opening books}
Randomized AIs can be seen as random samplers of deterministic policies.
A random seed is randomly drawn, and then a deterministic AI, depending on this seed, is applied. The choice of the random seed is usually considered of negligible importance. However, a recent work \cite{publinashrandomseed} has shown that random seeds have an impact, and that the bias inherent to the use of a given randomized AI, which has an implicit probability distribution on random seeds, can be significantly reduced by analyzing the impact of random seeds. We here extend this work to a more challenging case, namely Phantom Go.

Section \ref{pgo} describes Phantom Go, our testbed for experiments.
Section \ref{rsb} describes our approach for boosting random seeds.
Section \ref{xp} presents experimental results.

\section{Phantom Go}\label{pgo}
The game of Phantom Go is a two-player game with hidden information. It consists in playing Go without seeing the other player's moves. Each player does not see the board of the other player. In addition, there is a reference board that is managed by a referee and that the players do not see either. On each player's turn, the player proposes a move to the referee. If the move is legal on the reference board, it is played on the reference board and it is the other player's turn. If the move is illegal on the reference board, the referee tells the player that the move is illegal and the player is asked to play another move. The referee is in charge of maintaining the reference board and of telling illegal moves to the players. The game is over when the two players pass.

Monte Carlo methods have been used in Phantom Go since 2005. The resulting program plays at the level of strong human Go players. Monte Carlo Phantom Go was one of the early success of Monte Carlo methods in Go and related games. The principle of Monte Carlo Phantom Go is to randomly choose a ``determinization'' (i.e. a filling of the unknown parts of the state space) consistent with the previous illegal moves before each playout. For each possible move, the move is simulated, followed by a determinization and a random playout. Thousands of such determinizations and playout sequences are played for each move and the move with the highest resulting mean is played.

This simple method has defeated more elaborate methods using Monte Carlo Tree Search in the former computer Olympiads. Using a parallelization of the algorithm on a cluster our program won five gold medals and one silver medal during the last six computer Olympiads. When winning the silver medal, the program lost to another program using the same method. The program also played three strong Go players in exhibition matches during the 2011 European Go Congress and won all of its three games.

\section{Random seeds and their boosting}\label{rsb}
    \subsection{Seeds in games}\label{sec:prs}
We consider a randomized artificial intelligence (AI), equipped with random seeds. Our experiments will be performed on a Monte Carlo approach for Phantom Go, though the method is generic and could be applied to any randomized algorithm such as those cited in Section \ref{randai}.

We can decide the seed - and when the seed is fixed, the AI becomes deterministic. The original (randomized) algorithm can be seen as a probability distribution over these deterministic AIs.

A Random-Seed-portfolio (RS-portfolio) consists in optimizing the probability distribution on random seeds.
Such an algorithm has been proposed in \cite{publinashrandomseed}. We recall below the two algorithms they propose,
namely Nash and BestSeed. In both cases, the learning of the probability distribution is based on
the construction of a $K\times K$ binary matrix $M$, where $M_{i,j}=1$ if Black with random seed $i$ wins against White with random seed $j$, and $M_{i,j}=0$ otherwise. This matrix is the learning set; for validating our approach in terms of performance against the original randomized algorithm, we use random seeds which are not in this matrix, and distributed as in the original randomized algorithm.

    \subsection{Strategies for choosing seeds}
We describe here two methods for choosing a probability distribution on rows $i\in\{1,2,\dots,K\}$ and a probability distribution on columns $j\in \{1,2,\dots,K\}$. These probability distributions are then used as better probability distributions on random seeds at the beginning of later games.

  \subsubsection{BestSeed approach.}
{BestSeed is quite simple; the probability distribution for Black has mass $1$ on some $i$ such that $\sum_{j\in\{1,\dots,K\}} M_{i,j}$ is maximal.
We randomly break ties. For White, we have probability $1$ for some $j$ such that $\sum_{i\in \{1,\dots,K\}} M_{i,j}$ is minimum.}
The BestSeed approach is described in Algorithm \ref{bestseed}.
This method is quite simple, and works because
$$\lim_{K\to\infty} \frac1K \sum_{j=1}^K M_{i,j}$$
$$(resp. \ \lim_{K\to\infty} \frac1K \sum_{j=1}^K M_{j,i})$$
is far from being a constant when $i$ varies.

\begin{algorithm}[tbp]
\begin{algorithmic}[1]
\REQUIRE{$K$, and a randomized AI.}
\FOR{$i\in\{1,\dots,K\}$}
\FOR{$j\in\{1,\dots,K\}$}
\STATE{Play a game between
\begin{itemize}
\item an AI playing with seed $i$ as Black;
\item an AI playing with seed $j$ as White.
\end{itemize}
}
\STATE{$M_{i,j}\leftarrow1$ if Black wins, $0$ otherwise}
\ENDFOR
\ENDFOR
\STATE{$i_0 \leftarrow \argmax_{i \in \{1,\dots,K\}} \sum_{j=1}^K M_{i,j}$}
\STATE{$j_0 \leftarrow \argmin_{j \in \{1,\dots,K\}} \sum_{i=1}^K M_{i,j}$}
\RETURN{The (deterministic) AI using seed $i_0$ when playing Black and $j_0$ when playing White.}
\end{algorithmic}
\caption{\label{bestseed}The BestSeed algorithm for boosting a randomized AI. There is a parameter $K$; $K$ greater leads to better performance but slower computations. The resulting AI is deterministic, but it can be made stochastic by random permutations of the 8 symmetries of the board.}
\end{algorithm}

    \subsubsection{Nash approach.}
This section describes the Nash approach. It is more complicated than the BestSeed approach, but it is harder to \emph{exploit}, as detailed in the experimental section. First, we introduce constant-sum matrix games, and then we explain how we use them for building portfolios of random seeds.

\paragraph{Constant-sum matrix games}
We consider constant-sum matrix games; by normalizing matrices, we work without loss of generality on games such that the sum of the rewards for player 1 and for player 2 is one.
Consider the following game, parametrized by a $K\times K$ matrix $M$.
Black plays $i$. White is not informed of Black's choice, and plays $j$.
The reward for Black is $M_{i,j}$ and the reward for White is $1-M_{i,j}$.

It is known \cite{vonn,nash52} that there exists at least one Nash equilibrium $(x,y)$ such that if Black plays $i$ with probability $x_i$ and White plays $j$ with probability $y_j$, then neither of the players can improve its expected reward by changing unilaterally his policy. More formally:
$$ \exists (x,y),~\forall (x',y'),~x'^tMy\leq x^tMy\leq x^tMy', $$
where $x$, $y$, $x'$ and $y'$ are non-negative vectors summing to one.
Moreover, the value $v=x^tMy$ is unique - but the pair $(x,y)$ is not necessarily unique.

It is possible to compute $x$ and $y$ in polynomial time, using linear programming \cite{Stengel02computeequilibria}. Some faster methods provide approximate results in sublinear time \cite{grigoriadis,auer95gambling}. Importantly, these fast approximation algorithms are mathematically proved and do not require all the elements of the matrix to be available - only $O(K\log(K)/\e)$ elements in the matrix have to be computed for a fixed precision $\e>0$ on the Nash equilibrium.

\paragraph{Nash portfolio of random seeds}
Consider $(x,y)$ the Nash equilibrium of the matrix game $M$, obtained by e.g. linear programming.
Then the Nash method uses $x$ as a probability distribution over random seeds for Black and uses $y$ as a probability distribution over random seeds for White. The algorithm is detailed in Algorithm \ref{nash}.
\begin{algorithm}[tbp]
\begin{algorithmic}[1]
\REQUIRE{$K$ and a randomized AI.}
\FOR{$i\in\{1,\dots,K\}$}
\FOR{$j\in\{1,\dots,K\}$}
\STATE{Play a game between
\begin{itemize}
\item an AI playing with seed $i$ as Black;
\item an AI playing with seed $j$ as White.
\end{itemize}
}
\STATE{$M_{i,j}\leftarrow1$ if Black wins, $0$ otherwise}
\ENDFOR
\ENDFOR
\STATE{Let $(x,y)$ be a pair of probability distributions over $\{1,\dots,K\}$, forming a Nash equilibrium of $M$.}
\RETURN{The (stochastic) AI using seed
\begin{itemize}
\item $i_0$ randomly drawn with probability distribution $x$ when playing Black
\item and $j_0$ randomly drawn with probability distribution $y$ when playing White.\end{itemize}}
\end{algorithmic}
\caption{\label{nash}The Nash method for boosting a randomized AI. There is a parameter $K$; $K$ greater leads to better performance but slower computations. The resulting AI is stochastic. It is outperformed by BestSeed in terms of winning rate against the original (randomized) algorithm, but harder to overfit. }
\end{algorithm}
We also tested a sparse version, which gets rid of pure strategies with low values. The algorithm depends on a parameter $\alpha$, and it is detailed in Algorithm \ref{sparsenash}.
\begin{algorithm}[tbp]
\begin{algorithmic}[1]
\REQUIRE{$K$ $\alpha$ and a randomized AI.}
\FOR{$i\in\{1,\dots,K\}$}
\FOR{$j\in\{1,\dots,K\}$}
\STATE{Play a game between
\begin{itemize}
\item an AI playing with seed $i$ as Black;
\item an AI playing with seed $j$ as White.
\end{itemize}
}
\STATE{$M_{i,j}\leftarrow1$ if Black wins, $0$ otherwise}
\ENDFOR
\ENDFOR
\STATE{Let $(x,y)$ be a pair of probability distributions over $\{1,\dots,K\}$, forming a Nash equilibrium of $M$.}
\STATE{$x_{max}\leftarrow \max_{1\leq i\leq K} x_i$}
\STATE{$y_{max}\leftarrow \max_{1\leq i\leq K} y_i$}
\STATE{For all $i\in \{1,\dots,K\}$, if $x_i<\alpha x_{max}$, then $x_i\leftarrow 0$.}
\STATE{For all $i\in \{1,\dots,K\}$, if $y_i<\alpha y_{max}$, then $y_i\leftarrow 0$.}
\STATE{$x\leftarrow x/\sum_{i=1}^K x_i$}
\STATE{$y\leftarrow y/\sum_{i=1}^K y_i$}
\RETURN{The (stochastic) AI using seed
\begin{itemize}
\item $i_0$ randomly drawn with probability distribution $x$ when playing Black
\item and $j_0$ randomly drawn with probability distribution $y$ when playing White.\end{itemize}}
\end{algorithmic}
\caption{\label{sparsenash}The SparseNash method for boosting a randomized AI. Compared to Algorithm \ref{nash}, there is an additional parameter $\alpha$.}
\end{algorithm}
\begin{figure*}[t]
\centering
\includegraphics[width=0.32\linewidth]{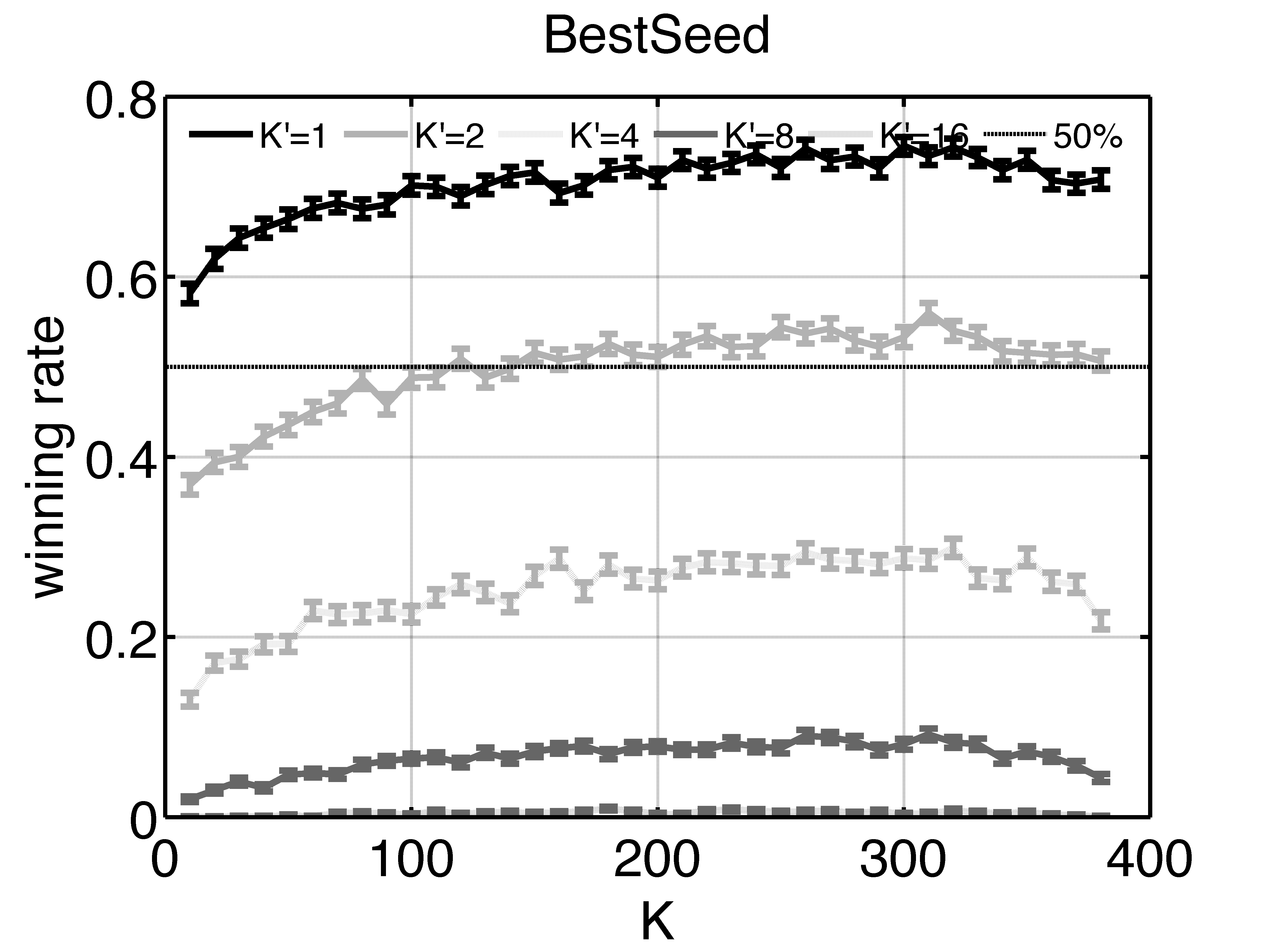}
\includegraphics[width=0.32\linewidth]{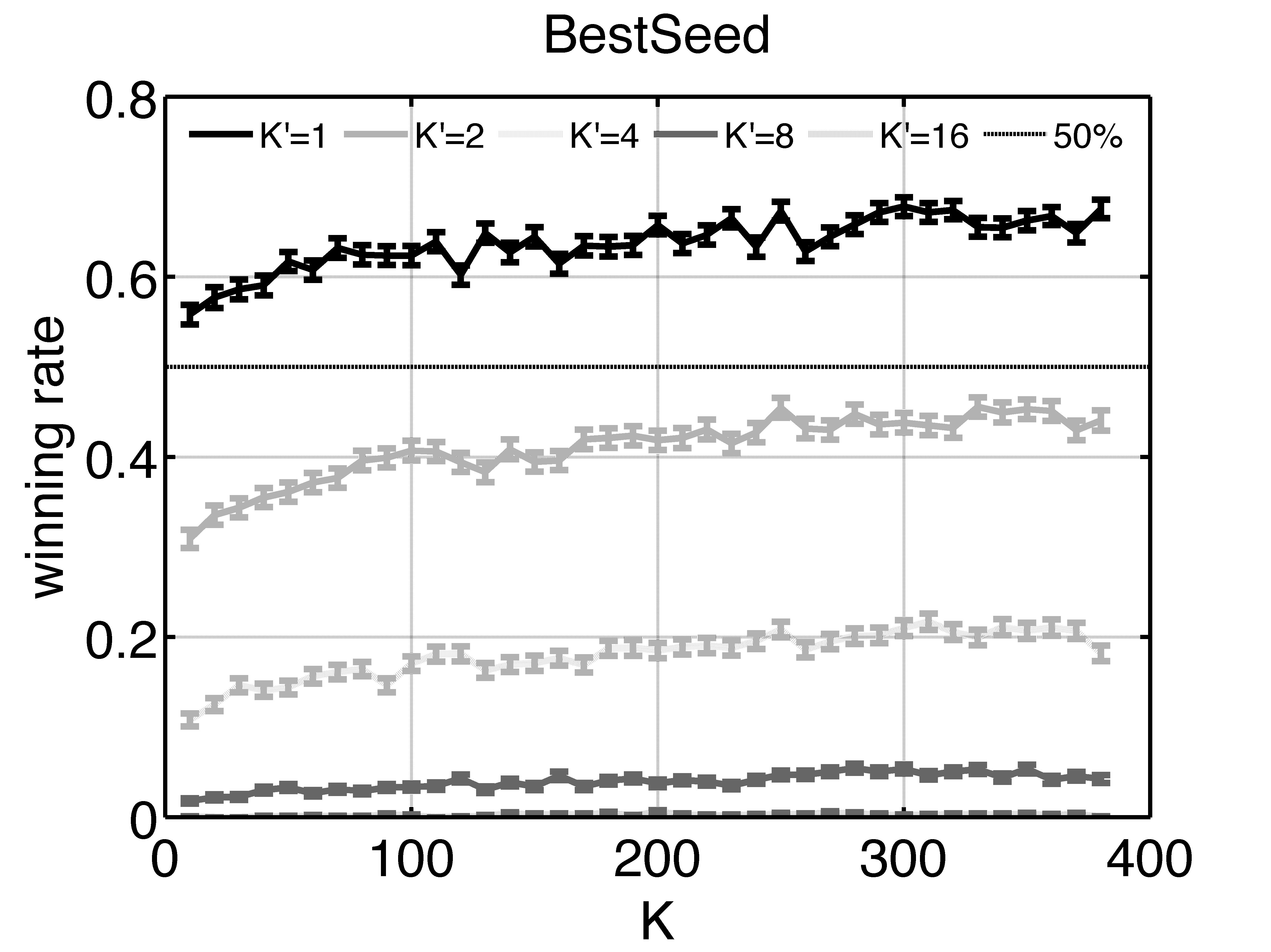}
\includegraphics[width=0.32\linewidth]{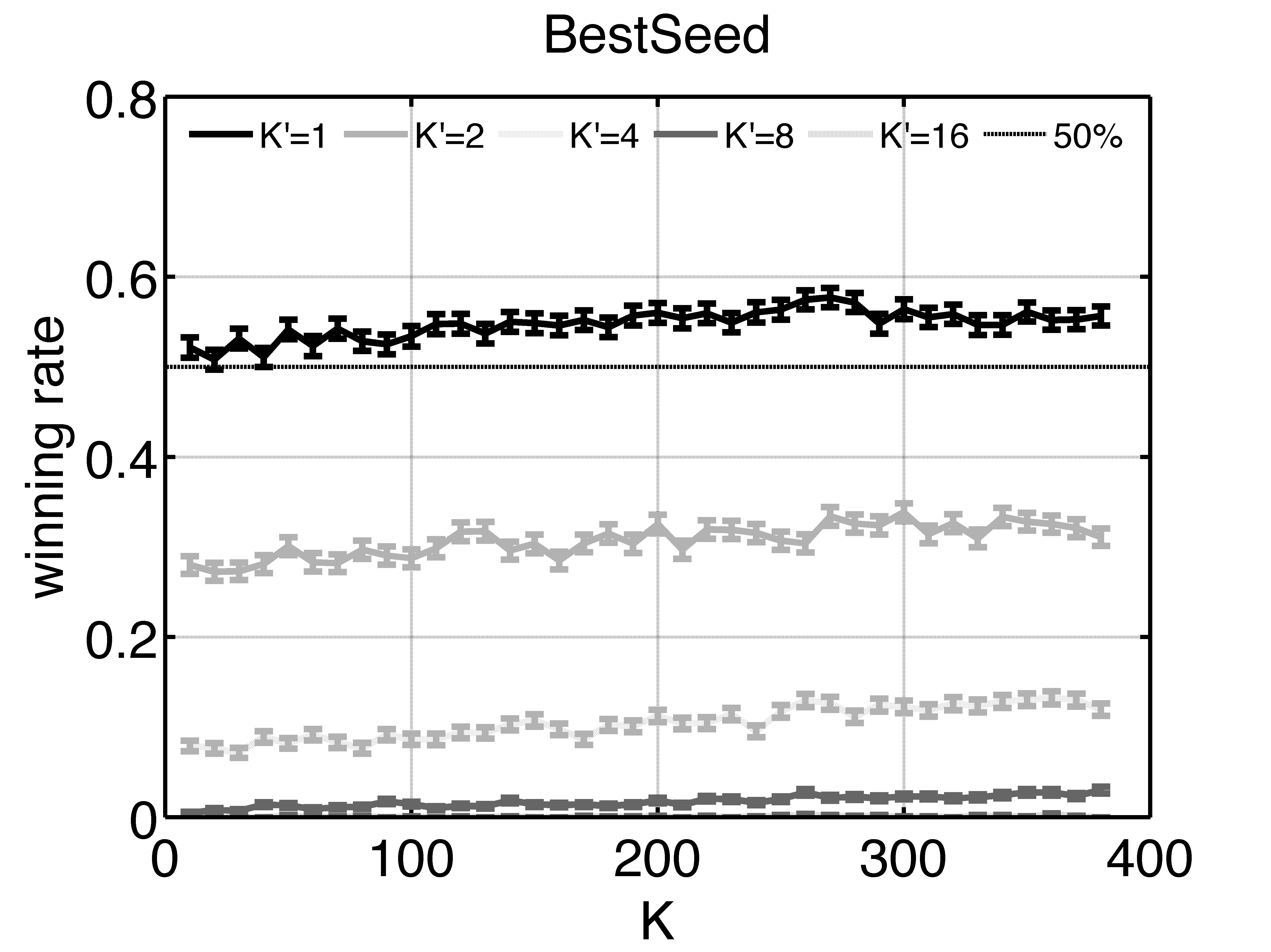}\\
\includegraphics[width=0.32\linewidth]{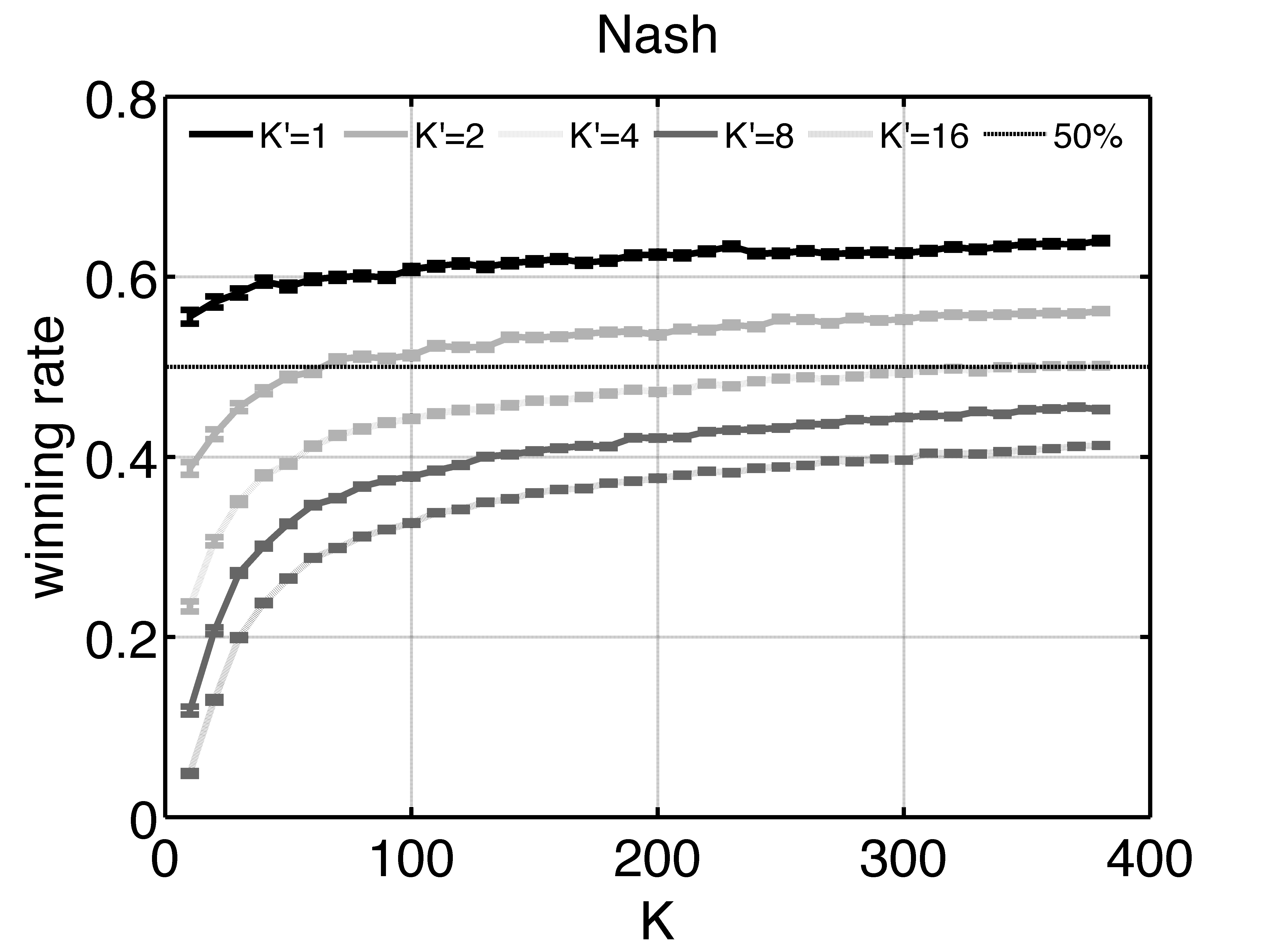}
\includegraphics[width=0.32\linewidth]{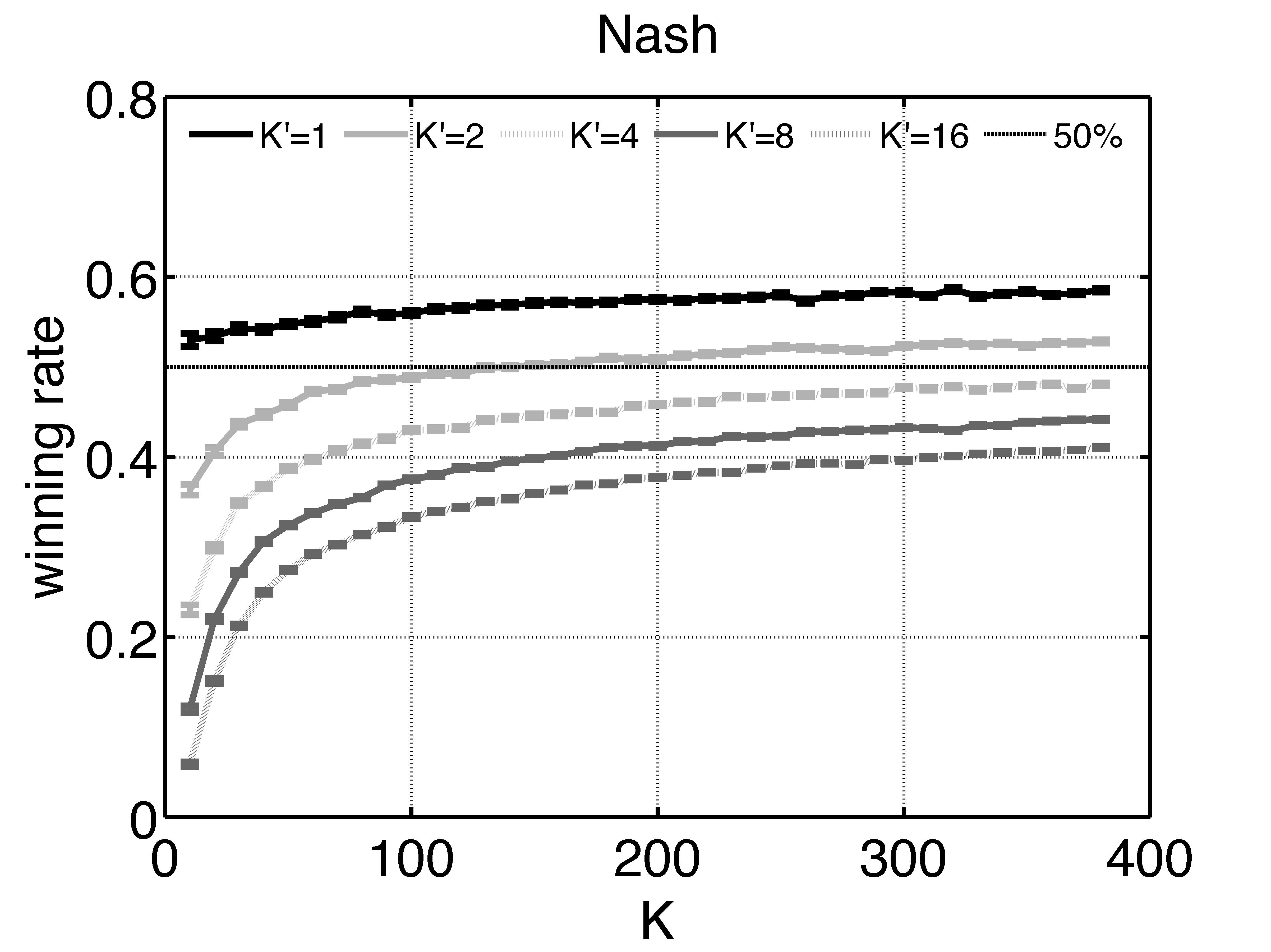}
\includegraphics[width=0.32\linewidth]{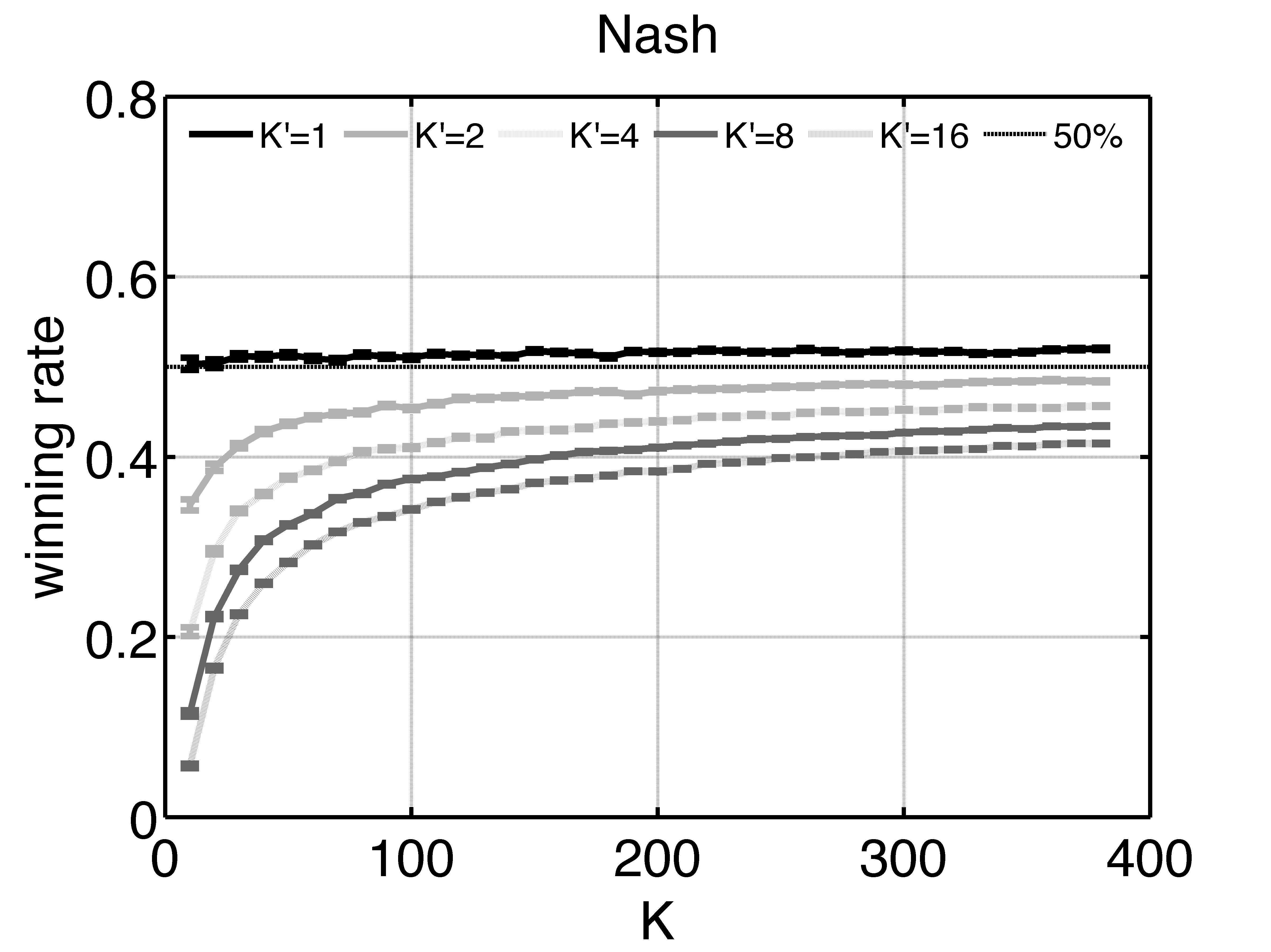}
\caption{\label{fig:board5}
Winning rate for Phantom Go 5x5 (left), 7x7 (middle) and 9x9 (right).
Top: BestSeed; bottom: Nash. X-axis: $K$ such that we learn on a $K\times K$ matrix $M$. Y-axis: winning rate.
The training is performed on a $K\times K$ matrix. The testing is performed on $K'\times K'$ strategies, i.e. $K'$ strategies for Black and $K'$ strategies for White. $K'=1$ corresponds to a randomized seed - this is therefore the original randomized AI, and performance greater than 50\% for $K'=1$ means that we outperformed the original randomized AI. $K'>1$ corresponds to the best performing AI, for each color, among $K'$ randomly drawn seeds; this is a very difficult opponent, who can try $K'^2$ strategies and keep only the best of their results.
The black dashed curve refers to a winning rate$=50\%$.
The experiments are repeated $1000$ times.
The standard deviations are shown in the plots.
}
\end{figure*}

    \subsection{Criteria}\label{criteria}
We now give two performance criteria, namely performance against the baseline (which is the original randomized algorithm) and performance against an agent which can choose its seed with perfect rationality among a finite randomly drawn set of a given cardinal - the second criterion is harder, and simulates an opponent who has ``learnt'' how to play against us by optimizing his seed. 

   	 \subsubsection{Performance against the baseline.}
The first criterion is the success rate against the original AI, with its randomized seed. This is precisely the criterion that is optimized in BestSeed; but we perform experiments in cross-validation, i.e. the performance obtained by BestSeed and displayed in Section \ref{xp} is the performance against seeds which were not used in the learning phase.

   	 \subsubsection{Performance against an opponent who learns.}\label{exploiter}
The second criterion is the success rate against an opponent who plays with the original randomized AI, but can test $K'$ randomly drawn random seeds and can select the best of these seeds.
This modelizes the case in which the opponent can choose (perfect choice) one policy among $K'$ policies. We here consider $K'$ policies, each of them obtained by fixing the random seed to some random value. We do this choice among $K'$ policies for Black, and $K'$ policies for White as well, so that we have indeed the worst performance against $K'\times K'$ policies. This becomes a very tough criterion when $K'$ increases - our opponent can basically test $(K')^2$ openings and choose the one for which we are the weakest.

Obviously, all experiments in the present paper are performed with separate seeds for the learning and the validation experiments, so that no overfitting can explain the positive results in Section \ref{xp}.

\section{Experimental results}\label{xp}
We perform experiments on the Phantom Go testbed. Our randomized AI is Golois \cite{phantomgo,phantomgo2}.
In all our results, we use cross-validation; we test performance against seeds which were not used during the learning phase. We consider values of $K\leq 400$. We use the two criteria described in Section \ref{criteria}, i.e. winning rate against the original randomized algorithm and worst of the winning rates against $K'\times K'$ deterministic policies obtained as described in Section \ref{exploiter}.
All presented winning rates are the average between the winning rate as Black and the winning rate as White.

\begin{table*}[t]
\centering
\begin{scriptsize}
\begin{tabular*}{\textwidth}{c@{\extracolsep{\fill}}cc@{\extracolsep{\fill}}c@{\extracolsep{\fill}}c@{\extracolsep{\fill}}c@{\extracolsep{\fill}}c@{\extracolsep{\fill}}c}
\hline
\multirow{2}{*}{Board} & \multicolumn{2}{c}{\multirow{2}{*}{Method}} & \multicolumn{5}{c}{Winning rate (\%)} \\
\cline{4-8}
 & && $K'=1$ & $K'=2$ & $K'=4$ & $K'=8$ & $K'=16$\\
\hline
\multirow{5}{*}{5x5} 
 & \multicolumn{2}{c}{Baseline} & $50 $ & $30.5 \pm 0.9$ & $12.5 \pm 0.7$ & $0.5 \pm 0.2$ & $0.0 \pm 0.0$\\
& \multicolumn{2}{c}{BestSeed} & $70.7 \pm 1.0$ & $49.8 \pm 1.1$ & $23.4 \pm 0.9$ & $4.8 \pm 0.4$ & $0.2 \pm 0.1$\\
& \multicolumn{2}{c}{Nash} & $63.8 \pm 0.3$ & $56.1 \pm 0.2$ & ${\bf{50.3 \pm 0.2}}$ & ${\bf{45.4 \pm 0.2}}$ & ${\bf{41.3 \pm 0.2}}$\\
&\multirow{3}{*}{Sparse}
 & $\alpha=0.500$ & $68.3 \pm 0.6$ & $56.4 \pm 0.6$ & $43.9 \pm 0.5$ & $32.8 \pm 0.4$ & $24.4 \pm 0.3$\\
& & $\alpha=0.750$ & ${\bf{74.7 \pm 0.8}}$ & ${\bf{57.7 \pm 0.9}}$ & $36.7 \pm 0.9$ & $20.2 \pm 0.6$ & $9.0 \pm 0.4$\\
& & $\alpha=1.000$ & $76.2 \pm 0.9$ & $55.2 \pm 1.1$ & $31.9 \pm 1.0$ & $8.7 \pm 0.6$ & $0.9 \pm 0.2$\\
 \hline
\multirow{5}{*}{7x7}
& \multicolumn{2}{c}{Baseline} & $50$ & $23.0 \pm 0.9$ & $8.5 \pm 0.6$ & $0.5 \pm 0.2$ & $0.5 \pm 0.2$\\
& \multicolumn{2}{c}{BestSeed} & ${\bf{66.5 \pm 1.0}}$ & $44.1 \pm 1.1$ & $19.9 \pm 0.9$ & $3.9 \pm 0.4$ & $0.1 \pm 0.1$\\
& \multicolumn{2}{c}{Nash} & $58.3 \pm 0.2$ & ${\bf{52.8 \pm 0.2}}$ & ${\bf{48.1 \pm 0.2}}$ & ${\bf{44.2 \pm 0.1}}$ & ${\bf{41.1 \pm 0.1}}$\\
&\multirow{3}{*}{Sparse}
 & $\alpha=0.500$ & $59.6 \pm 0.3$ & $51.1 \pm 0.3$ & $44.0 \pm 0.2$ & $38.7 \pm 0.2$ & $33.2 \pm 0.2$\\
& & $\alpha=0.750$ & $58.7 \pm 0.6$ & $44.1 \pm 0.6$ & $30.9 \pm 0.5$ & $21.4 \pm 0.4$ & $14.5 \pm 0.3$\\
& & $\alpha=1.000$ & $56.4 \pm 1.1$ & $33.0 \pm 1.0$ & $13.7 \pm 0.8$ & $1.7 \pm 0.3$ & $0.0 \pm 0.0$\\
 \hline
\multirow{5}{*}{9x9}
& \multicolumn{2}{c}{Baseline} & $50$ & $27.0 \pm 1.0$ & $4.0 \pm 0.4$ & $1.0 \pm 0.2$ & $0.0 \pm 0.0$\\
& \multicolumn{2}{c}{BestSeed} & ${\bf{54.4 \pm 1.1}}$ & $32.8 \pm 1.0$ & $12.2 \pm 0.7$ & $2.8 \pm 0.4$ & $0.1 \pm 0.0$\\
& \multicolumn{2}{c}{Nash} & $51.9 \pm 0.1$ & ${\bf{48.4 \pm 0.1}}$ & ${\bf{45.6 \pm 0.1}}$ & ${\bf{43.5 \pm 0.1}}$ & ${\bf{41.6 \pm 0.1}}$\\
&\multirow{3}{*}{Sparse}
 & $\alpha=0.500$ & $52.2 \pm 0.3$ & $45.3 \pm 0.2$ & $39.4 \pm 0.2$ & $35.3 \pm 0.2$ & $31.1 \pm 0.2$\\
& & $\alpha=0.750$ & $52.4 \pm 0.6$ & $38.6 \pm 0.5$ & $27.6 \pm 0.4$ & $18.4 \pm 0.4$ & $12.5 \pm 0.3$\\
& & $\alpha=1.000$ & $52.9 \pm 1.1$ & $27.3 \pm 1.0$ & $8.2 \pm 0.6$ & $1.2 \pm 0.2$ & $0.1 \pm 0.1$\\
\hline
\end{tabular*}
\end{scriptsize}
\caption{\label{tab:rate}Winning rate for Phantom Go 5x5, 7x7 and 9x9 with $K=380$ (cf. Figure \ref{fig:board5}).
$\alpha$ is the sparsity parameter (cf. Algorithm \ref{sparsenash}).
The experiments are repeated $1000$ times.
The standard deviations are shown after $\pm$.
$K'=1$ corresponds to the original algorithm with randomized seed; $K'=2$ corresponds to the original algorithm but choosing optimally (after checking their performance against its opponent) between 2 possible seeds, i.e. it is guessing, in an omniscient manner, between 2 seeds, each time an opponent is provided. $K'=4$, $K'=8$, $K'=16$ are similar with 4, 8, 16 seeds respectively; $K'=16$ is a very strong opponent for our original algorithm (our winning rate is initially close to 0), but after Nash seed learning we get results above 40\% in 5x5, 7x7 and 9x9.
}
\end{table*}

\begin{figure*}[t]
\centering
\includegraphics[width=0.32\linewidth]{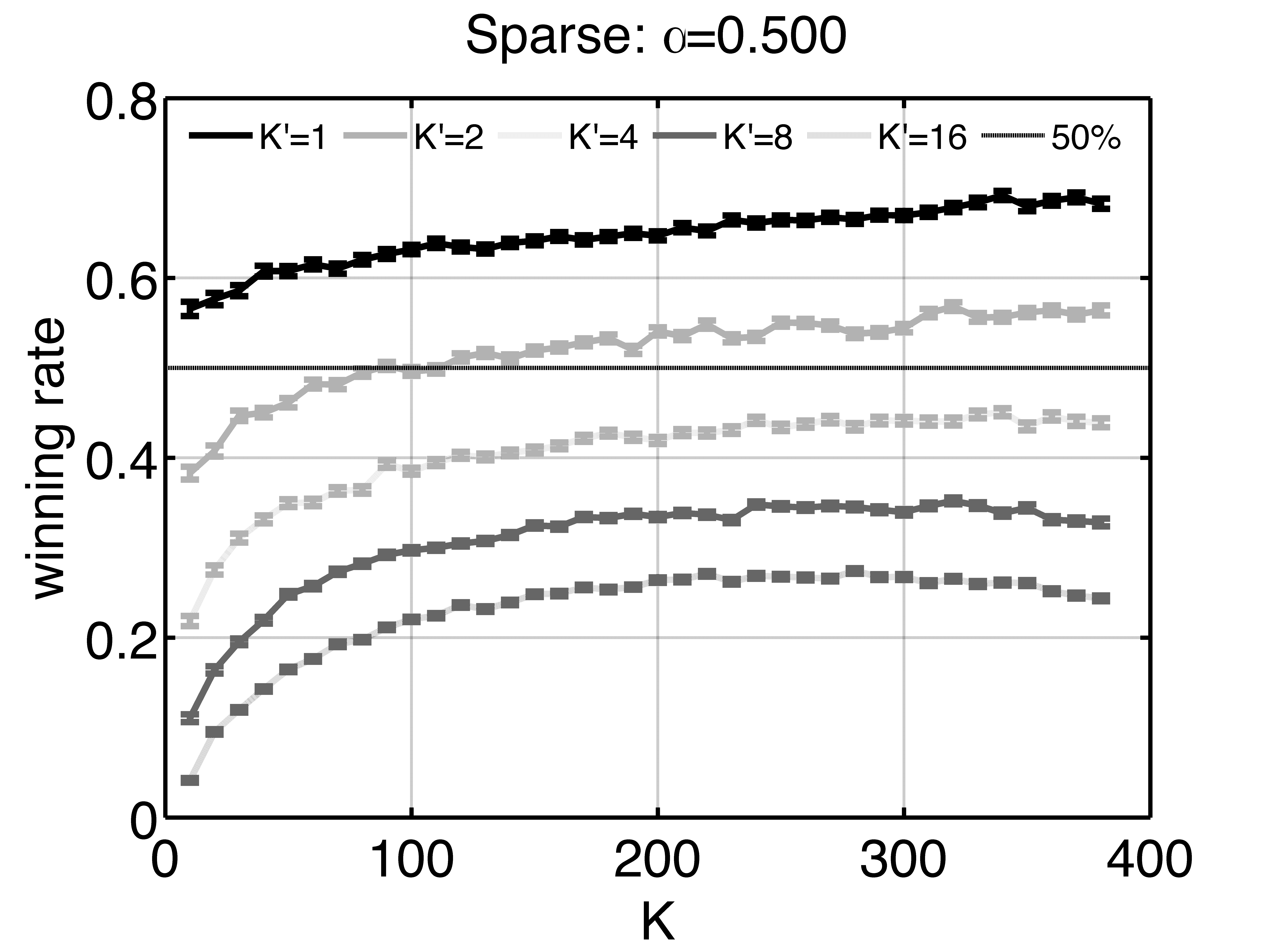}
\includegraphics[width=0.32\linewidth]{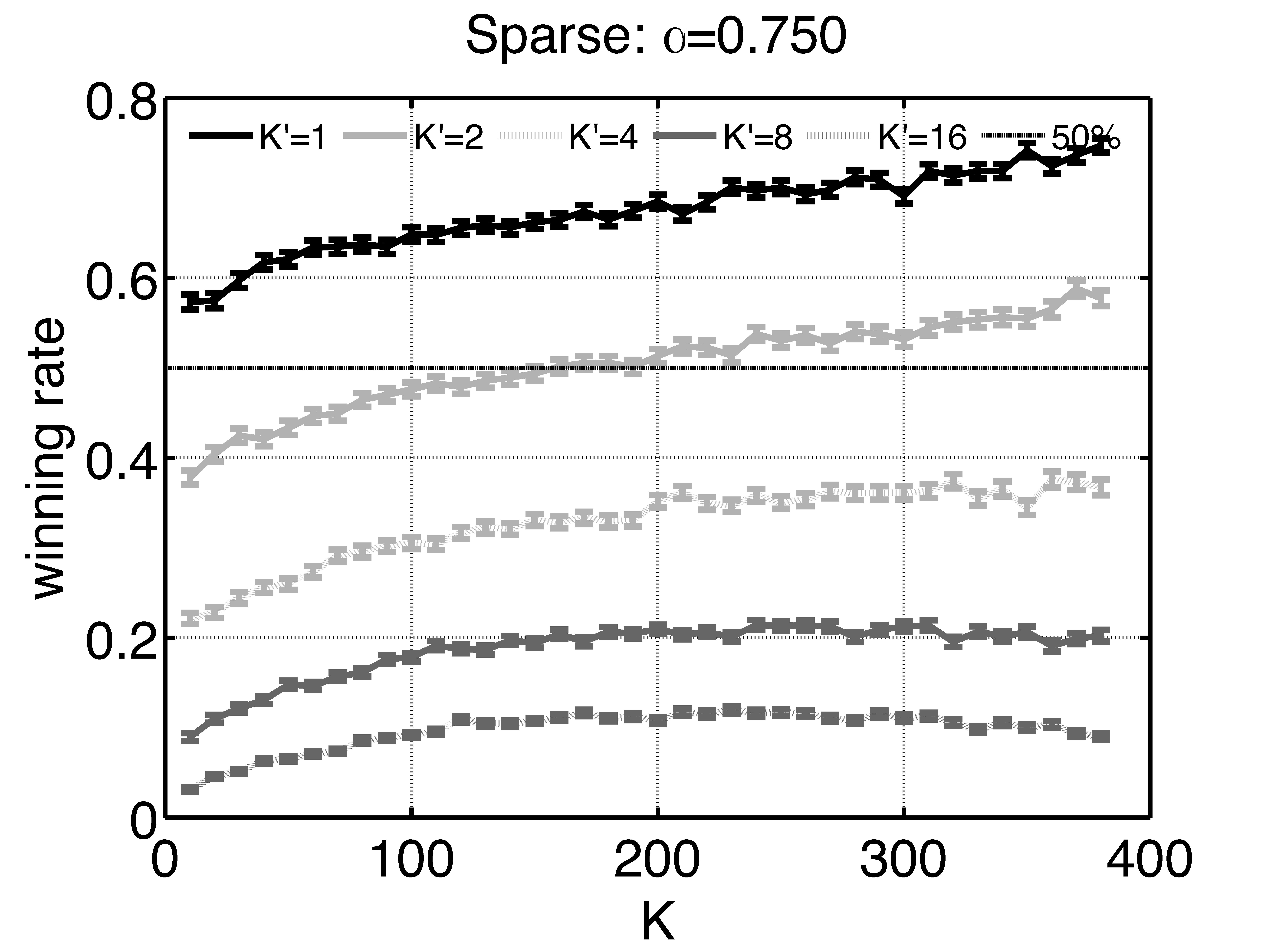}
\includegraphics[width=0.32\linewidth]{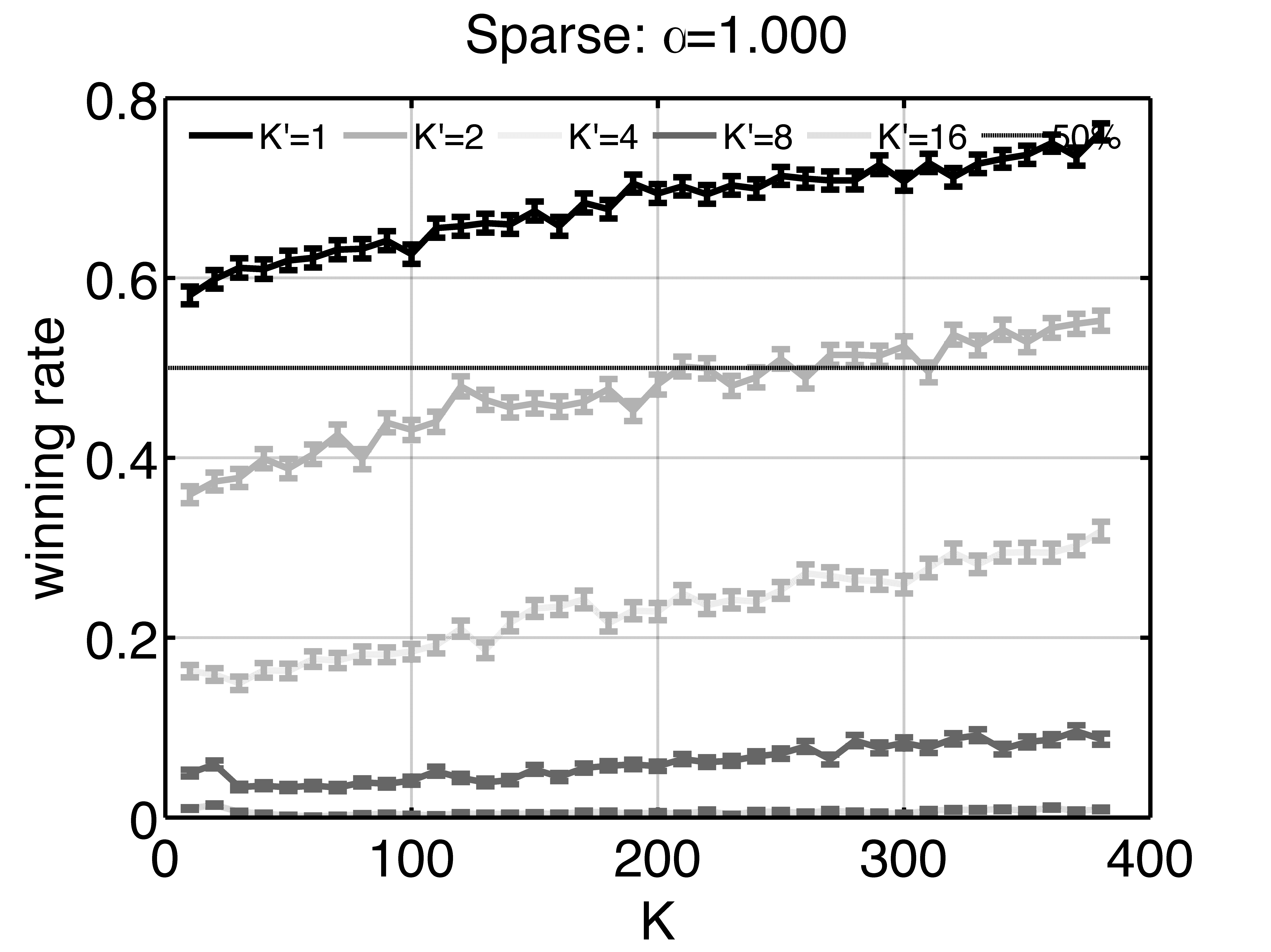}\\
\includegraphics[width=0.32\linewidth]{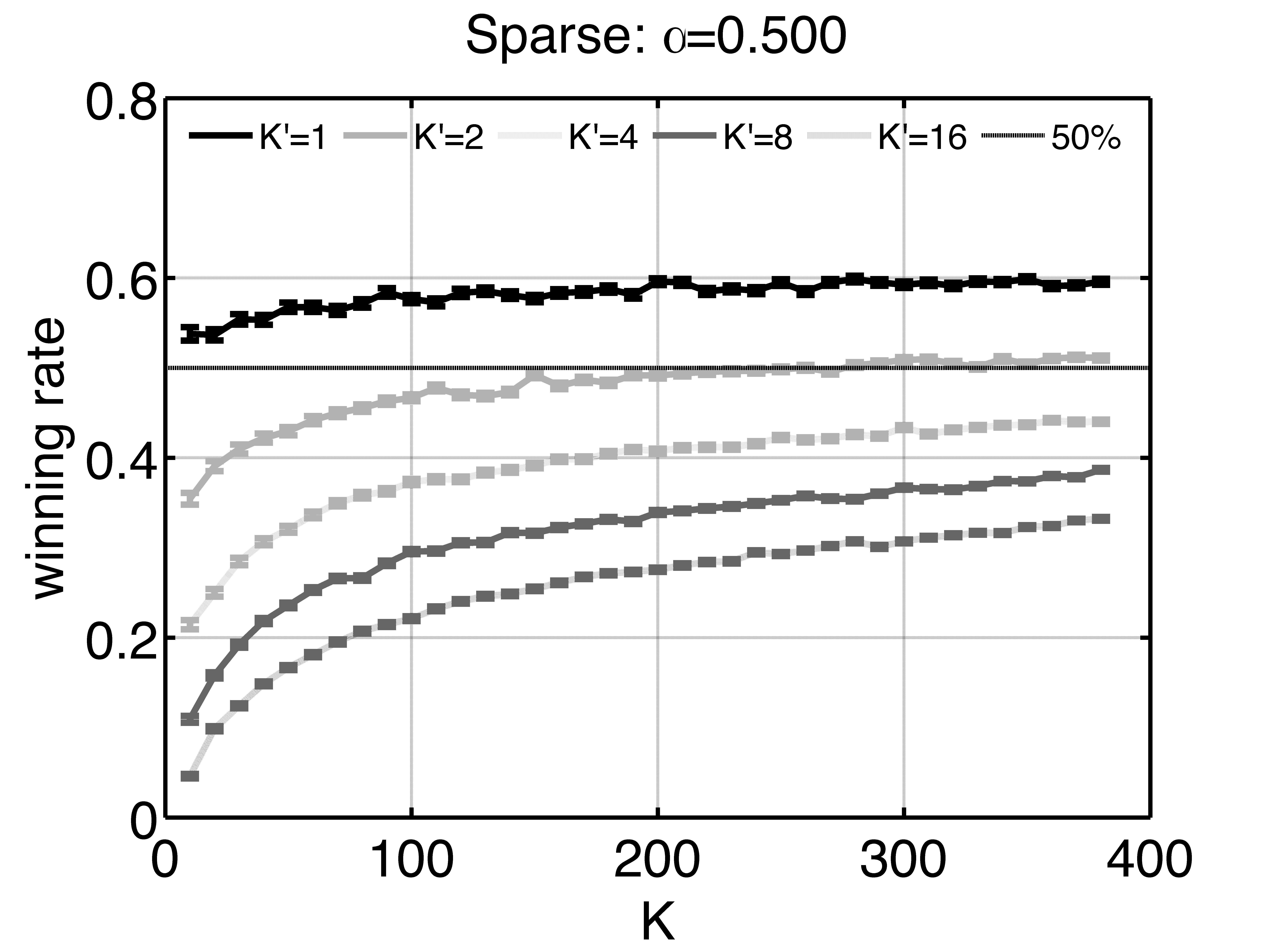}
\includegraphics[width=0.32\linewidth]{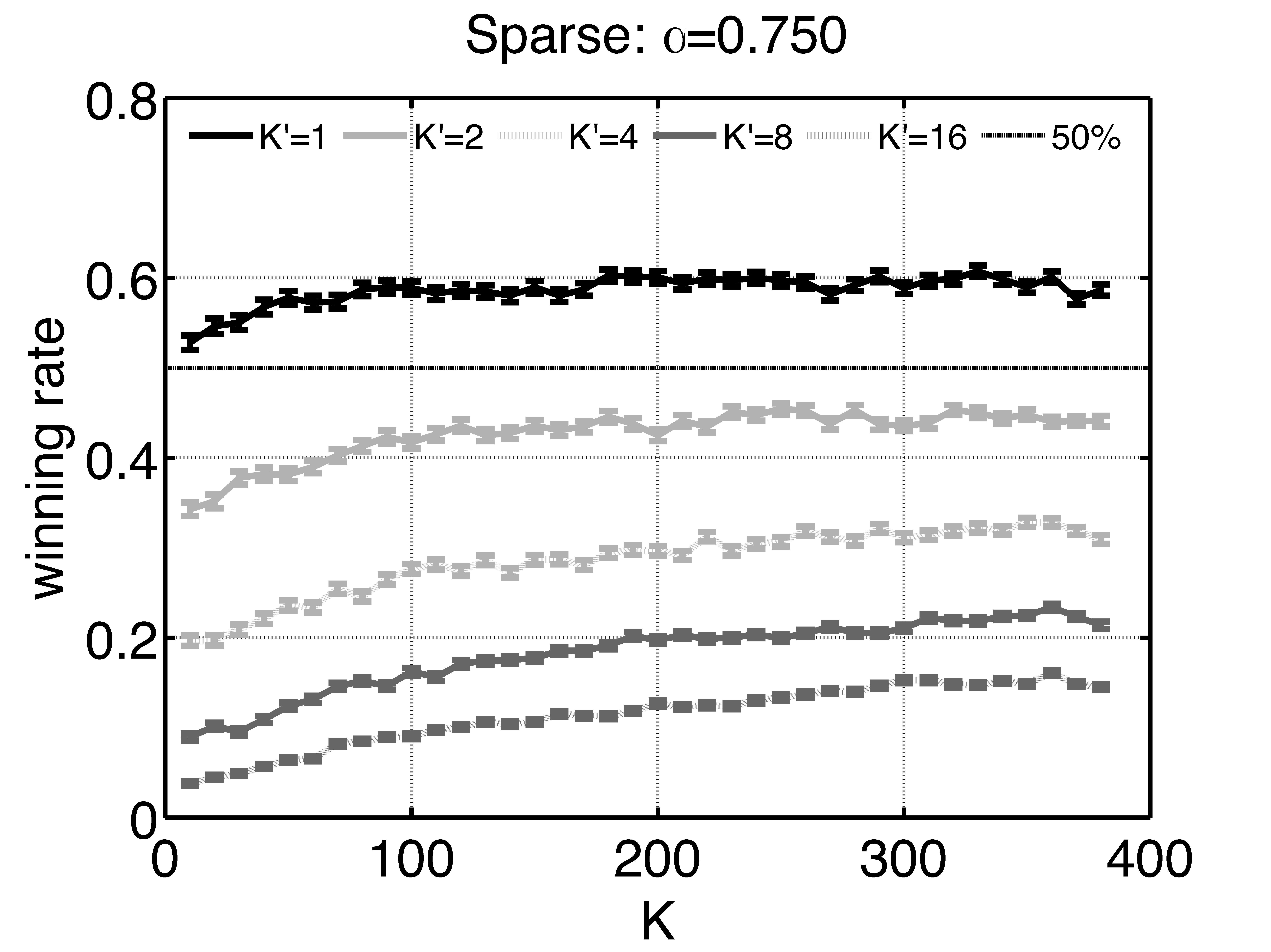}
\includegraphics[width=0.32\linewidth]{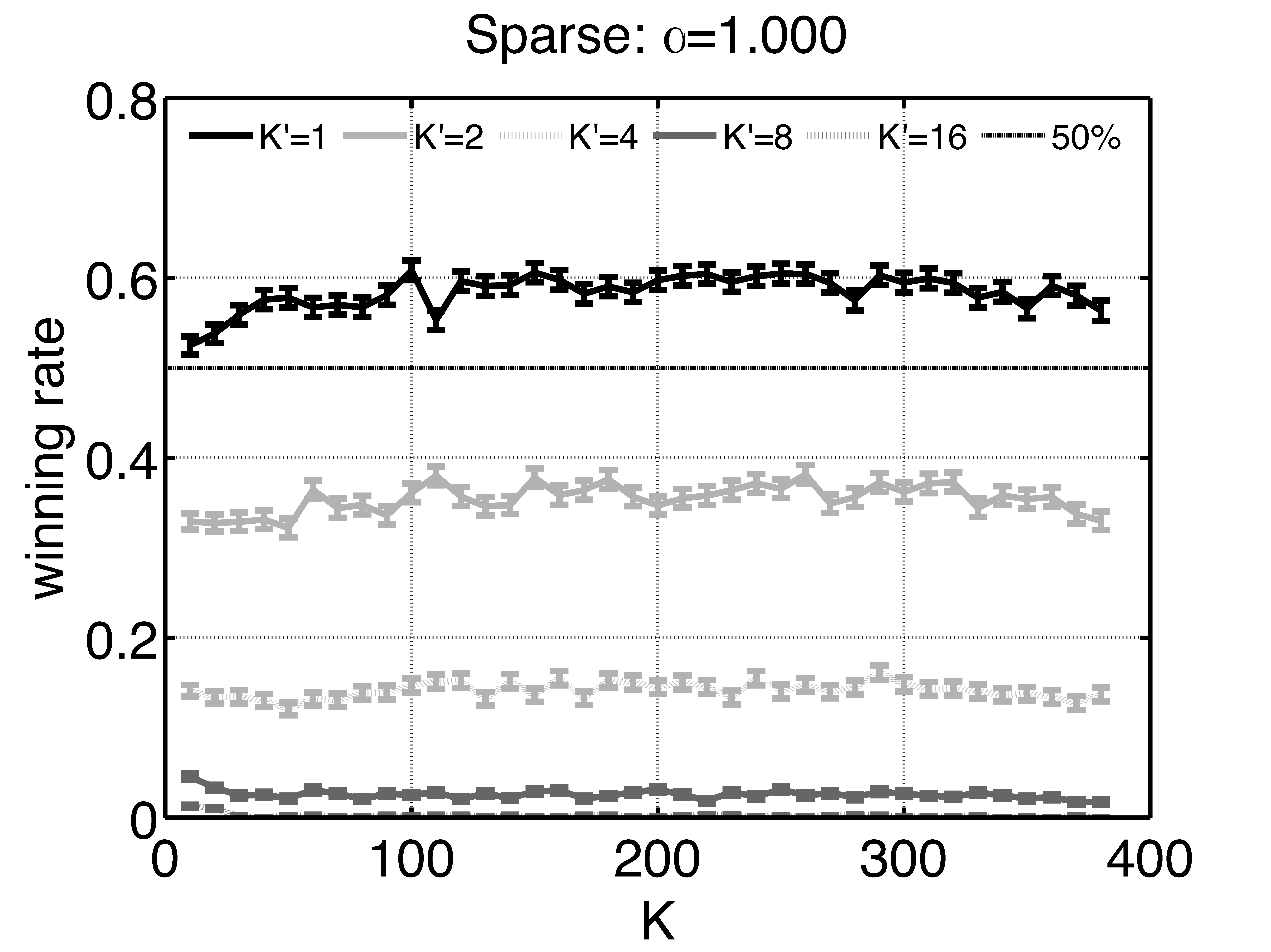}\\
\includegraphics[width=0.33\linewidth]{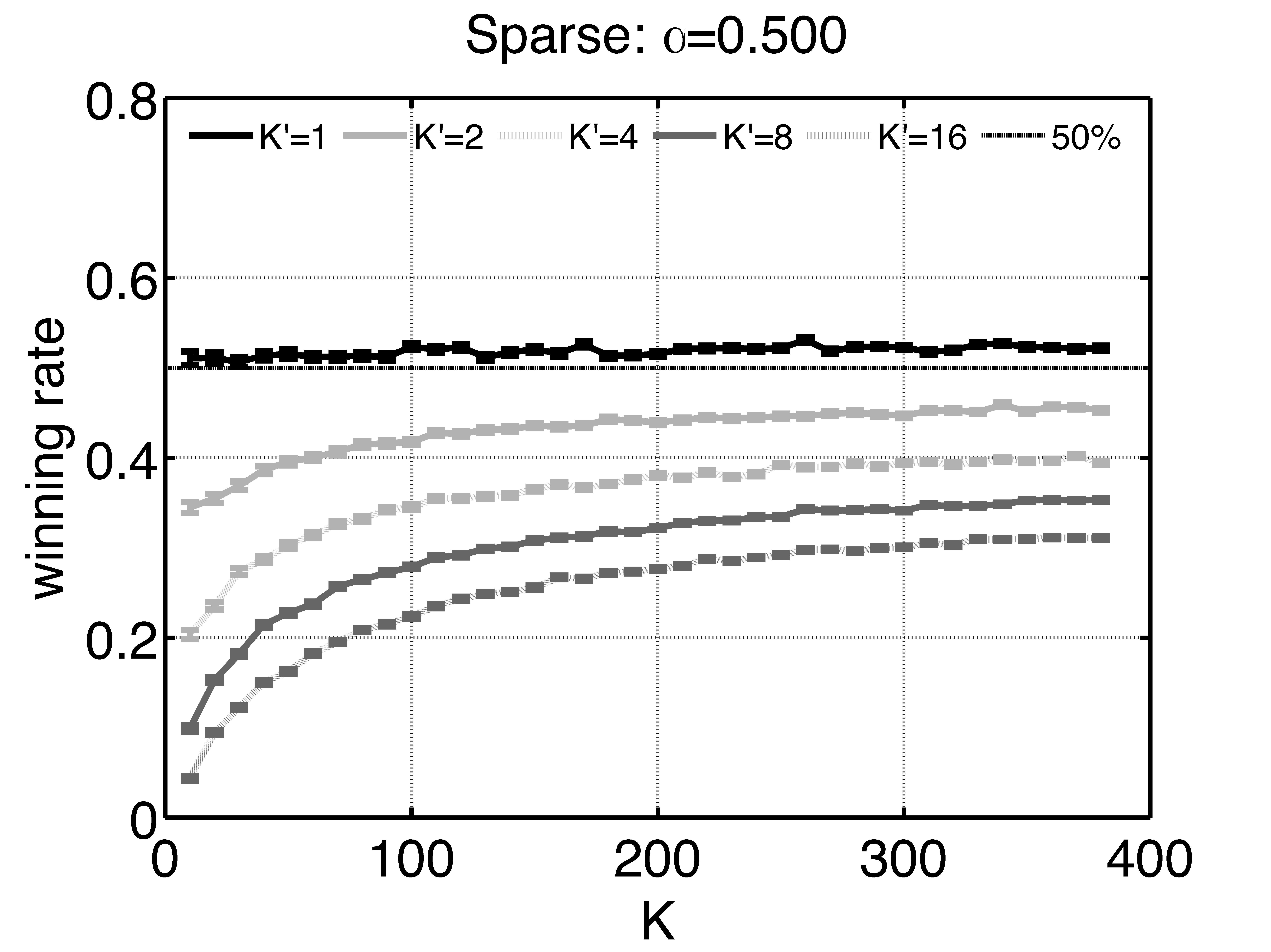}
\includegraphics[width=0.33\linewidth]{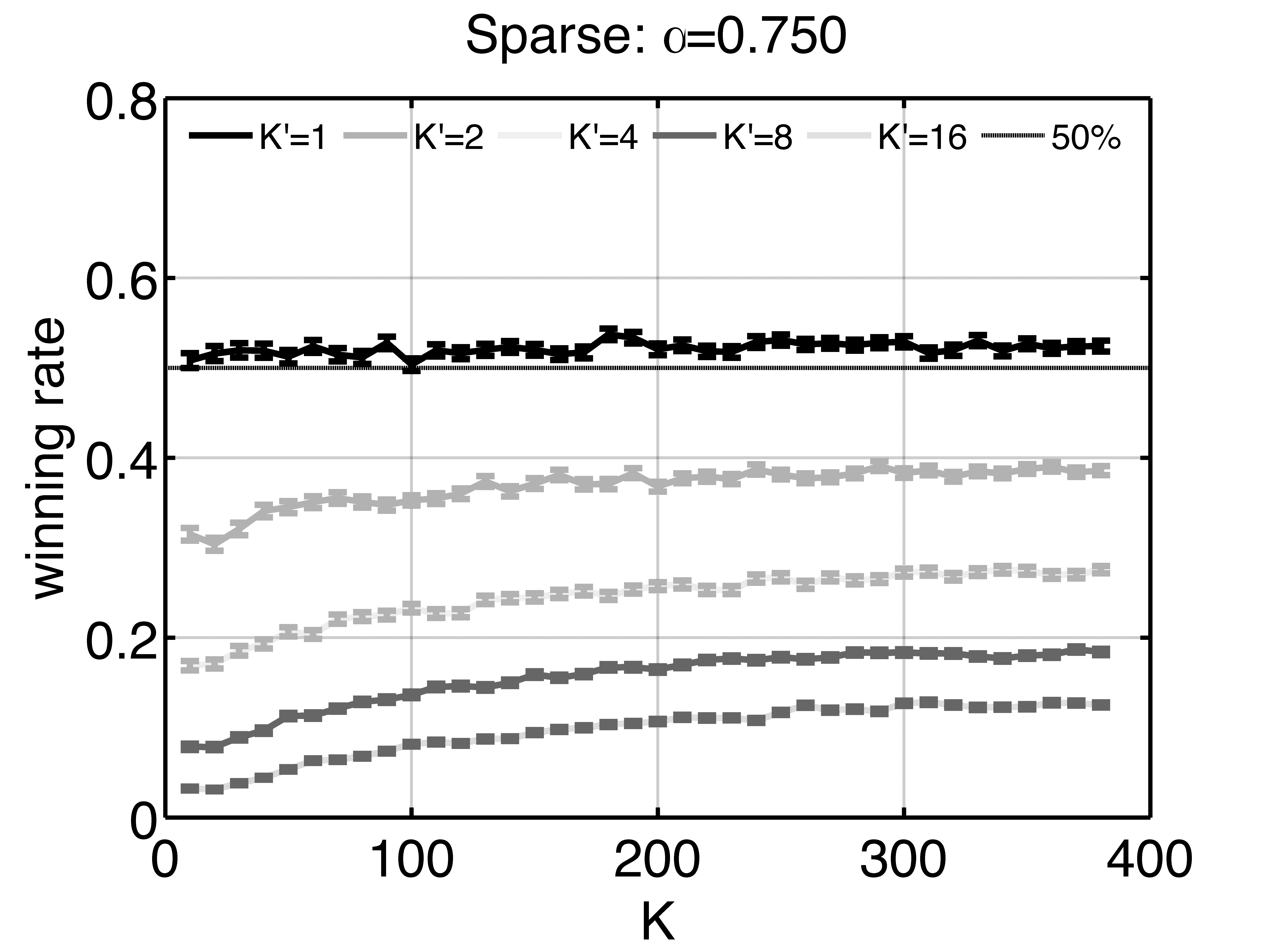}
\includegraphics[width=0.33\linewidth]{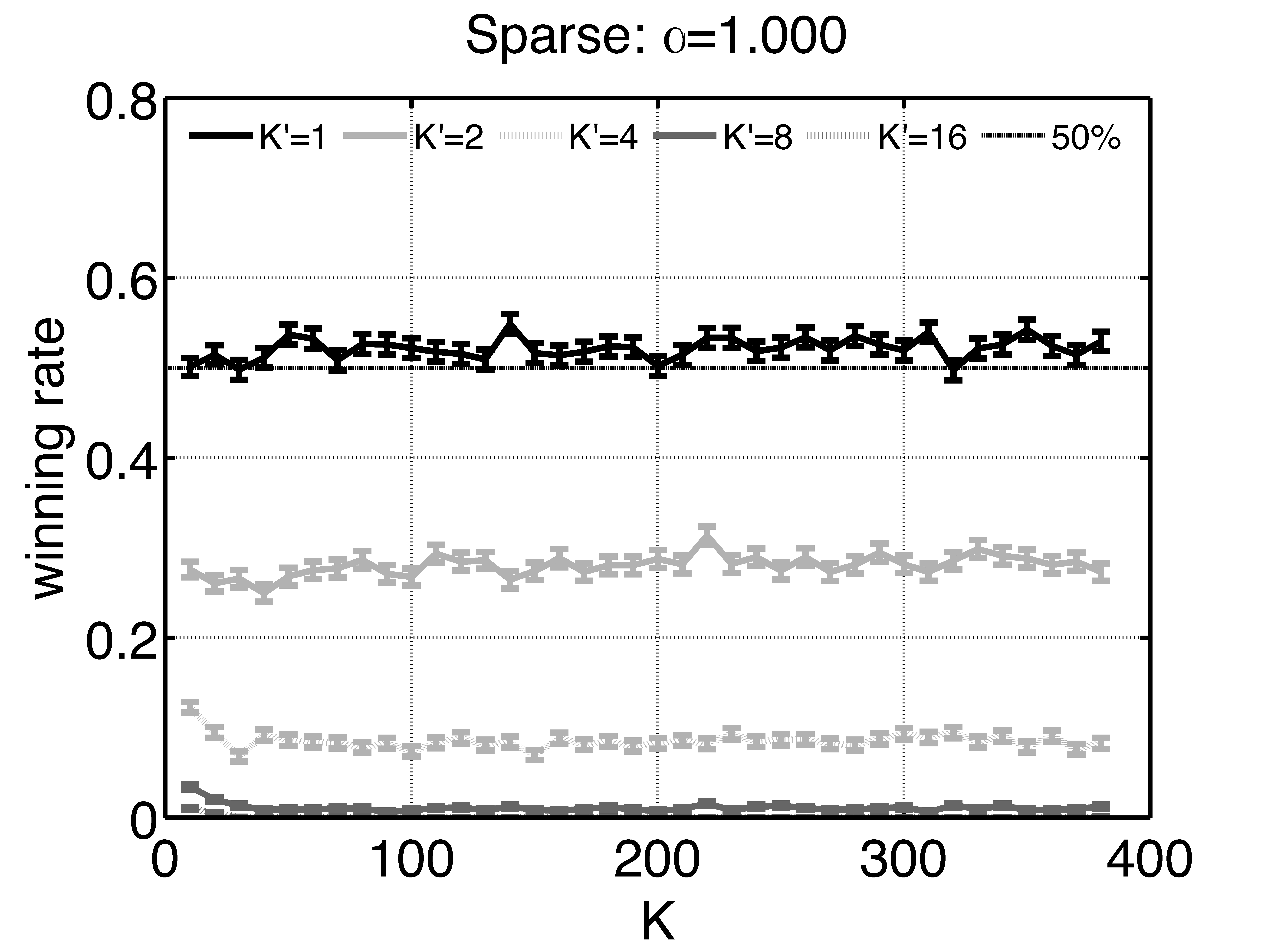}
\caption{\label{fig:sparse579}Winning rate for Phantom Go 5x5 (top), 7x7 (middle) and 9x9 (bottom) using sparse strategy with different sparsity parameter $\alpha$. X-axis: parameter $K$ (size of the learning set). Y-axis: performance in generalization against the original algorithm ($K'=1$) and against the learning opponent (see Section \ref{exploiter}; $K'=2$ to $K'=16$).}
\end{figure*}

We observe in Figure \ref{fig:board5} (also Table \ref{tab:rate}):
\begin{itemize}
{
    \item The BestSeed approach clearly outperforms the original method in 5x5, 7x7 and 9x9. The performance is excellent in 5x5, greater than 71\%; around 67\% in 7x7; it is still good in 9x9 (54\%).
    \item The Nash approach reaches 64\% in 5x5, 58\% in 7x7. This is already reasonably good for a very randomized game such as Phantom Go; in partially observable games like Phantom Go, Poker, or many card games, several games are usually required for knowing the best among two players. In 9x9, we got only 52\% - not very impressive.
    \item The SparseNash approach outperforms BestSeed in terms of success rate against the original randomized AI, in 5x5 (Figure \ref{fig:sparse579} (top), summarized in Table \ref{tab:rate}). Results are however disappointing on larger board sizes (Figure \ref{fig:sparse579} (middle and bottom); also presented in Table \ref{tab:rate}). 
}
\end{itemize}
These two methods were tested directly on the original algorithm, without using the symmetries of the game or any prior knowledge. All results are obtained with proper cross-validation. Standard deviations are shown on figures and are negligible compared to deviations from 50\%.
The approach has a significant offline computational cost; but the online computational overhead is zero. The offline computational overhead is $K^2$ times the cost of one game, plus the Nash solving. The Nash solving by linear programming is negligible in our experiments. For large scale matrices, methods such as \cite{grigoriadis} should provide much faster results as the number of games would be $O(K\log(K)/\e^2)$ instead of $K^2$ for a fixed precision $\e$.

\section{Conclusions}
We tested various methods for enhancing randomized AIs by optimizing the probability distribution on random seeds. 
Some of our methods are not new, but up to now, they were only tested on a fully observable game, without opening book, whereas in fully observable games building an opening book is far less a challenge. 
We work on Phantom Go, a very challenging problem, with the program which won most competitions in recent years. The three tested methods provide results as follows:
\begin{itemize}
\item With BestSeed, we get 71\%, 67\%, 54\% of success rate against the baseline in 5x5, 7x7 and 9x9, just by ``managing'' the seeds. 
\item The Nash approach provides interesting results as well, in particular strongly boosting the performance against stronger opponent such as $K'=2$, $K'=4$, $K'=8$, $K'=16$, reaching 40\% (in 5x5, 7x7 and 9x9) whereas our original algorithm was close to 0\% winning rate for $K'=16$. This means that the opening book we have learnt is robust against stronger opponents than the ones used for the self-play involved in our learning. 
\item Using the Nash approach with sparsity, with the exponent $\alpha=.75$ recommended in earlier papers on sparsity \cite{auger2014sparse}, maybe not the best for each case separately, but outperforming the baseline in all cases.
\end{itemize}
The method has no computational overhead online - all the computational cost is an offline learning. As a consequence, the method looks like a free bonus: when your randomized AI is ready, apply Algorithm \ref{nash} and get a better AI.
The BestSeed method is the best performing one, but it can be overfitted. The Nash approach is less efficient against the original AI, but more robust, i.e. more difficult to overfit.

\subsection*{Further work}
We propose the following further works:
\begin{itemize}
\item The approach is quite generic, and could be tested on many games in which randomized AIs are available. For the BestSeed approach, the game does not have to be a two-player game.
\item Our work does not use any of the natural symmetries of the game; this should be a very simple solution for greatly improving the results; in particular, it would be much harder to overfit BestSeed if it was randomized by the 8 classical board symmetries.
\item Mathematically analyzing the approach is difficult, because we have no assumption on the probability distribution of $\E_j M_{i,j}$ for a randomly drawn seed $i$ - how many $i$ should we test before we have a good probability of having a really good one? Bernstein inequalities \cite{Ber46,hoe63,chern} for the BestSeed approach, and classical properties of Nash equilibria for the Nash approach, provide only preliminary elements.
\item Computing approximate Nash equilibria (using \cite{grigoriadis} or \cite{auer95gambling}) should strongly reduce the offline computational cost. The computational cost was not a big deal for the results presented in the present paper, but performance might be much better with $K$ larger. Approximate Nash equilibria do not need the entire $K\times K$ matrix; they only sample $O(K\log(K)/\e^2)$ elements of the matrix for a precision $\e$.
\item This last further work opens some problems also in the algorithmic theory of Nash equilibria. We have done the present work in a not anytime manner; we know $K$ a priori, and we do not have any approximate results until the $K^2$ games are played. However, we might prefer not to choose a priori a number $K$ of games, and get anytime approximate results. To the best of our knowledge, \cite{grigoriadis,auer95gambling} have never been adapted to an infinite set of arms. Also, adversarial bandit approaches such as Exp3 \cite{auer95gambling} have never been parallelized. \cite{grigoriadis} is parallel, but possibly harder to adapt in an anytime setting.
\end{itemize}

\bibliography{../../sandraswork/Others/tout,../../sandraswork/Others/tout2,../../sandraswork/Others/tout3,bibNash,../../sandraswork/Others/teytaud,phantomgo}

\end{document}